\documentclass[letterpaper]{article} %
\usepackage{aaai2026}  %
\usepackage{times}  %
\usepackage{helvet}  %
\usepackage{courier}  %
\usepackage[hyphens]{url}  %
\usepackage{graphicx} %
\usepackage{natbib}  %
\usepackage{caption} %
\usepackage{algorithm}
\usepackage{algorithmic}

\usepackage{newfloat}
\usepackage{listings}
\DeclareCaptionStyle{ruled}{labelfont=normalfont,labelsep=colon,strut=off} %
\floatstyle{ruled}
\newfloat{listing}{tb}{lst}{}
\floatname{listing}{Listing}
\usepackage{microtype}

\title{AI-Driven Marine Robotics:\\Emerging Trends in Underwater Perception and Ecosystem Monitoring}
\author{
    Scarlett Raine, Tobias Fischer
}
\affiliations{
    QUT Centre for Robotics, Queensland University of Technology\\

    2 George St, Brisbane, QLD, Australia\\
    \{sg.raine, tobias.fischer\}@qut.edu.au
}

\usepackage{eso-pic}
\usepackage{url}
\AddToShipoutPicture*{%
     \AtTextUpperLeft{%
         \put(-3.5,10){
           \begin{minipage}{\textwidth}
              \scriptsize
              \MakeUppercase{PREPRINT OF AAAI Conference on Artificial Intelligence 2026 ARTICLE; final version available at:} \url{https://doi.org/10.1609/aaai.v40i48.42133}
           \end{minipage}}%
     }%
}

\begin{document}

\maketitle

\begin{abstract}
Marine ecosystems face increasing pressure due to climate change, driving the need for scalable, AI-powered monitoring solutions to inform effective conservation and restoration efforts. This paper examines the rapid emergence of underwater AI as a major research frontier and analyzes the factors that have transformed marine perception from a niche application into a catalyst for AI innovation. We identify three convergent drivers: i) environmental necessity for ecosystem-scale monitoring, ii) democratization of underwater datasets through citizen science platforms, and iii) researcher migration from saturated terrestrial computer vision domains. Our analysis reveals how unique underwater challenges—turbidity, cryptic species detection, expert annotation bottlenecks, and cross-ecosystem generalization—are driving fundamental advances in weakly supervised learning, open-set recognition, and robust perception under degraded conditions. We survey emerging trends in datasets, scene understanding and 3D reconstruction, highlighting the paradigm shift from passive observation toward AI-driven, targeted intervention capabilities. The paper demonstrates how underwater constraints are pushing the boundaries of foundation models, self-supervised learning, and perception, with methodological innovations that extend far beyond marine applications to benefit general computer vision, robotics, and environmental monitoring.

\end{abstract}

\section{Introduction}
\label{sec:intro}
Marine ecosystems, including coral reefs and seagrass meadows, are under increasing threat from climate change, with rising sea temperatures, ocean acidification, and more frequent extreme weather events driving widespread degradation~\cite{anthony2020interventions}. Effective monitoring and large-scale restoration are urgently needed to safeguard the biodiversity and ecological services these systems provide~\cite{ditria2022artificial}. However, the scale of the challenge is immense: coral reefs span hundreds of thousands of square kilometers globally, while seagrass meadows form vast and often remote coastal habitats that are difficult to access and monitor comprehensively~\cite{lavery2013variability}.

\begin{figure}[t]
    \centering
    \includegraphics[width=\linewidth,clip,trim=0cm 1.9cm 0cm 1cm]{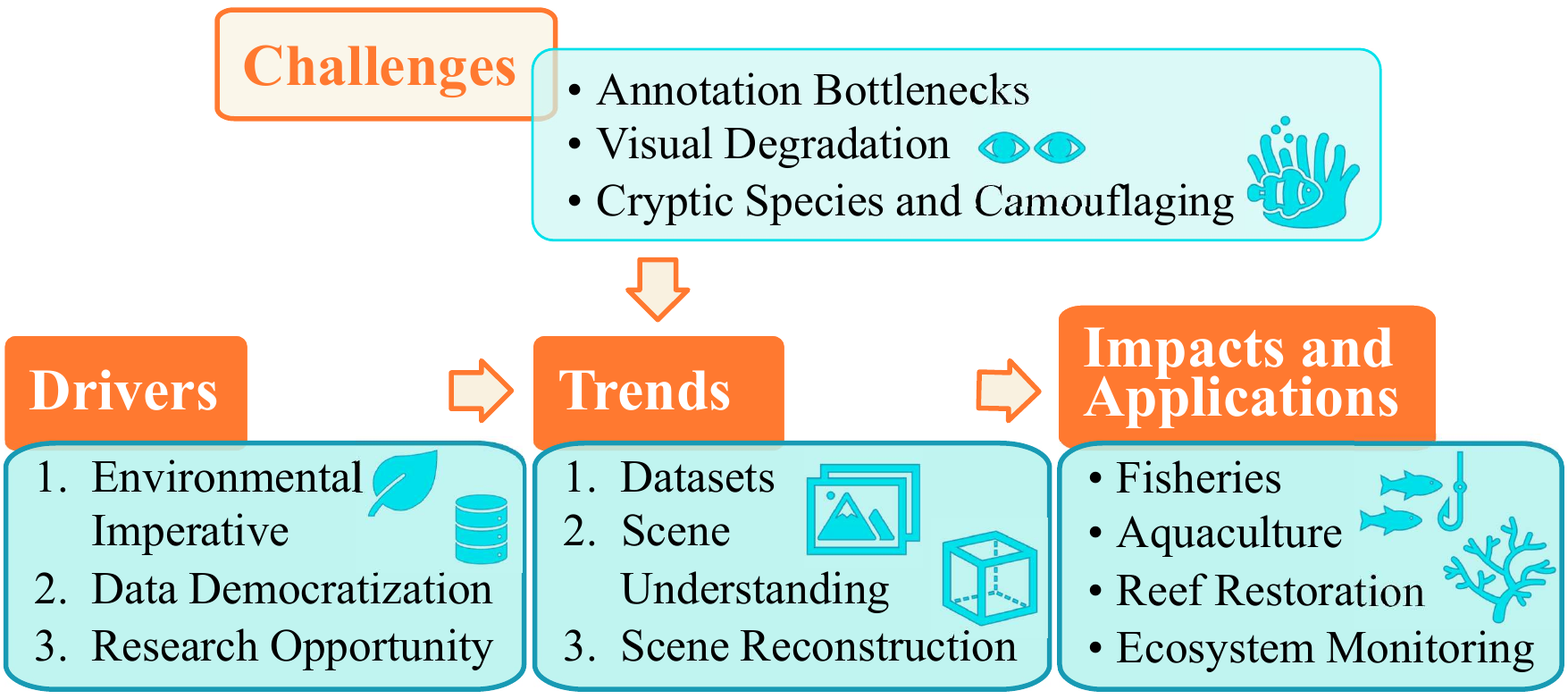}
    \caption{Visual outline.  We discuss the three main drivers for underwater AI as a research field of growing importance, identify key trends within underwater perception, placing these in the context of the underwater-specific challenges that have inspired innovations, and finally discuss the real-world impacts and applications of these advances.}
    \label{fig:frontpage}
\end{figure}

Traditional monitoring approaches, based on in-water surveys conducted by divers and domain experts, have played a central role in ecological assessment for decades~\cite{murphy2010observational}. Yet these methods are costly, time-intensive, and constrained by human safety limitations, dive duration, and accessibility to remote locations~\cite{mckenzie2020global, raine2020multi}. The deployment of robotic underwater and surface vehicles, along with towed camera platforms, has begun to mitigate these limitations by enabling the collection of large quantities of image data across previously inaccessible areas. However, this technological shift has created a new bottleneck: the need for automated approaches to analyze the vast quantities of visual data generated~\cite{gonzalez2020monitoring, raine2024reducing}.

This bottleneck has catalyzed the emergence of AI-driven underwater perception as a rapidly growing research frontier. The convergence of three key drivers is transforming what was once a niche application domain into a frontier for AI innovation (Fig.~\ref{fig:frontpage}). First, the environmental imperative for scalable marine monitoring has created unprecedented demand for automated solutions that can operate at ecosystem scales. Second, the democratization of underwater data through citizen science initiatives and the release of large-scale datasets has provided the data foundation required for modern AI. Third, as traditional computer vision applications become increasingly saturated, researchers are turning to the underwater domain as a comparatively understudied environment that offers opportunities for novel methodological contributions while addressing pressing societal needs.

These converging factors have enabled a paradigm shift in underwater perception: from manual species counts and expert-driven habitat assessments toward AI systems capable of real-time scene understanding, automated species detection, and even active intervention for habitat restoration. The unique challenges of the underwater environment, including turbidity, variable lighting conditions, cryptic species, and the need for fine-grained biological expertise, are driving innovations in self-supervised learning, open-set recognition, and robust perception that extend far beyond marine applications.

This paper provides a critical analysis of the emerging trends in AI for underwater perception and marine robotics (Fig.~\ref{fig:frontpage}). We first identify and motivate the key drivers of underwater perception as an emerging field (Section~\ref{sec:drivers}). 
Next, we discuss recent breakthroughs in key technical areas including datasets, scene understanding and 3D reconstruction (Section~\ref{sec:trends}), and conclude by exploring the broader implications of this emerging field for both marine science and the advancement of AI (Section~\ref{sec:conclusion}).

\section{Why Now? Convergent Drivers}
\label{sec:drivers}
The emergence of AI for underwater perception and marine robotics as a major research frontier is not coincidental, but rather the result of three powerful drivers converging at this moment in time. Understanding these drivers is crucial for appreciating both the urgency and the unprecedented opportunities that define this emerging field.

\subsection{Environmental Imperative}
The scale and urgency of marine ecosystem degradation has reached a tipping point, with some projections estimating a loss of up to 90\% of coral coverage before 2050~\cite{Souter2021Status}.  Ocean warming has accelerated dramatically, with marine heatwaves becoming five times more frequent since the 1980s, triggering mass coral bleaching events that can affect thousands of square kilometers~\cite{anthony2020interventions, henley2024highest}. In the face of this crisis, marine surveys are needed to detect changes early, track ecosystem health over time, and inform management and intervention strategies~\cite{ditria2022artificial}.  However, Australia's Great Barrier Reef alone spans over 344,000 square kilometers, an area larger than the entire United Kingdom, making comprehensive manual monitoring logistically impossible.

This crisis has created unprecedented demand for ecosystem-scale monitoring with minimal human intervention. Blue carbon markets and restoration programs likewise require scalable, quantitative assessments to support policy, carbon verification, and intervention tracking~\cite{gonzalez2020monitoring, murphy2010observational}. The gap between traditional dive-based surveys and the needs of large-scale conservation is now a major driver for AI-powered automation.

\subsection{Data Democratization}

The second driver represents a fundamental shift in how underwater data is collected, labeled, and shared. Historically, underwater imagery was manually collected by divers and snorkelers during expensive research expeditions and labeled by marine biologists, resulting in small, specialized datasets accessible only to domain experts. This paradigm has been disrupted by the convergence of several technological and social trends.

Underwater data collection has shifted from small, expert-driven efforts to large-scale, democratized initiatives powered by affordable imaging, citizen science, and corporate participation. Programs like Virtual Reef Diver, FathomVerse, and the Great Reef Census engage thousands of volunteers in annotation, while platforms such as CoralNet, ReefCloud, and Data Mermaid support semi-automated labeling. Combined with self-supervised learning, these developments create a feedback loop that accelerates data generation, model training, and public engagement (see Section~\ref{sec:trends}).

\subsection{Research Opportunity}
The third driver reflects a broader shift in the AI landscape. Traditional computer vision tasks reached a level of maturity where incremental improvements require vast computational resources and engineering effort. In these fields, foundation models are increasingly performing multiple computer vision tasks simultaneously~\cite{simeoni2025dinov3}.  Meanwhile, performance on standard benchmarks has plateaued, and many researchers are seeking new domains where fundamental algorithmic innovations can still yield transformative advances.

The underwater domain offers exactly this opportunity. Unlike terrestrial environments, underwater perception presents unique challenges that cannot be solved by simply applying existing methods: extreme lighting variations, backscatter and turbidity effects, color distortion with depth, and a diversity of marine life that exhibits camouflaging and cryptic behaviors. Foundation models developed for land-based applications typically do not directly translate to underwater applications, due to the sizable domain gap, largely caused by challenging underwater image characteristics, fine-grained classes, and ecological considerations~\cite{raine2024human}, as explained further in Section~\ref{sec:trends}. These challenges require genuine methodological innovation rather than incremental engineering improvements.

\subsection{Convergence Effect}

The true power of these drivers lies not in their individual impact, but in their convergence. Environmental urgency provides the motivation and funding, democratized data provides the foundation for modern AI approaches, and the research opportunity attracts top talent and computational resources. The underwater domain offers a compelling value proposition: the chance to develop novel AI techniques while addressing urgent environmental challenges. This has already attracted talent and funding from conservation agencies, creating a self-reinforcing cycle where advances enable new applications, generate more data, and drive further breakthroughs.

This convergence has reached a critical mass, transforming underwater perception from a niche application area into one of the most rapidly growing frontiers in AI research, with implications that extend far beyond marine science into fundamental questions of robust perception, domain adaptation, and human-AI collaboration in challenging environments.

\section{Emerging Trends}
\label{sec:trends}

This section provides a critical analysis of the recent and emerging trends in underwater perception.  We do not provide an exhaustive survey of prior approaches, but instead highlight key works that we believe are driving innovation in the field and opening up new research directions. Where relevant, we also identify the specific challenges of the underwater domain driving the technical advances.

\subsection{Datasets}
\label{subsec:datasets}

As mentioned in Section~\ref{sec:drivers}, the transformation of underwater AI has been fundamentally enabled by a dramatic shift in data availability, from small, expert-curated datasets to large-scale, community-driven repositories that rival terrestrial computer vision benchmarks. This data revolution represents more than simple scale increases; it reflects a fundamental change in how underwater imagery is collected, annotated, and shared across research communities.

\subsubsection{Citizen Science as a Data Engine:}
The proliferation of affordable underwater cameras, remotely operated vehicles (ROVs), and autonomous underwater vehicles (AUVs) has dramatically increased the volume of imagery being collected, enabling the emergence of large-scale, open-access datasets. Citizen science has become a powerful mechanism for annotating this data, democratizing participation in marine science at unprecedented scales.

One early initiative, the~\citet{virtualreefdiver}, encouraged divers and snorkelers to upload reef images, which were then annotated by everyday citizens. These observations, combined with professional monitoring data, improved predictive maps of coral cover across the Great Barrier Reef~\cite{vercelloni2023virtual}. Similarly, Citizens of the Reef has shown that citizen scientists can reach expert-level accuracy in coral reef monitoring, with deep learning and crowd-sourced analysis producing 99\% accurate benthic cover estimates across 8,086 images~\cite{lawson2025broadscale}. The Seatizen Atlas extends this model globally, contributing over 1.6 million geolocated images from the Southwest Indian Ocean collected by both citizens and researchers~\cite{contini2025seatizen}.  

\subsubsection{Gamification and Corporate Social Responsibility:}
The rise of Corporate Social Responsibility (CSR) initiatives and gamification has further driven public participation in labeling underwater imagery. Gamified platforms such as \citet{fathomverse}, NASA’s Nemo-Net~\cite{nemonet, van2021nemo}, and NOAA’s Click-a-Coral~\cite{clickacoral}, each with thousands of downloads and users, transform annotation tasks into engaging, interactive experiences. These projects make large-scale annotation efforts more achievable by reaching audiences beyond scientific communities.  

Complementing these gamified initiatives, the \citet{greatreefcensus} Great Reef Census integrates citizen science into education and corporate structures. The Schools Program allows students to learn about reef ecosystems while contributing data, and the Hour of Power initiative mobilizes CSR by encouraging employees from companies including Atlassian, Dell, Disney, Mars, Cotton On, and Salesforce to contribute one hour of labor for labeling reef imagery. Together, gamification and CSR expand participation, enhance ocean literacy, and strengthen stewardship of marine ecosystems.  

\subsubsection{Specialized Datasets Driving Technical Innovation:}
Specialized datasets are also pushing the boundaries of what underwater AI can achieve. CoralVOS introduces dense coral video segmentation, enabling frame-by-frame tracking of coral organisms and advancing beyond sparse point-based analysis~\cite{ziqiang2023coralvos}. The Sweet Corals dataset demonstrates the integration of 3D reconstruction and species classification, providing more than 90,000 georeferenced images alongside 3D point clouds and Gaussian Splatting models that enable both spatial and biological analysis~\cite{wildflow_sweet_corals_indo_2025}. CoralScapes establishes the first general-purpose dense semantic segmentation benchmark for coral reefs, analogous to Cityscapes in terrestrial computer vision~\cite{sauder2025coralscapes}.  FathomNet represents a paradigm shift in marine data curation, providing expert-annotated imagery spanning the full diversity of ocean life, from common coastal species to rarely encountered deep-sea organisms that present novel vision challenges~\cite{katija2022fathomnet}. 

\subsubsection{Data Consolidation and Standardization:}
Various efforts consolidate fragmented datasets collected under different labeling schemes and methodologies, unifying resources across diverse seafloor environments. With more than 11.4 million curated images and 3.1 million annotations standardized to consistent taxonomic schemes, BenthicNet enables large-scale model training that generalizes across ecosystems and imaging conditions~\cite{lowe2025benthicnet}.  MarineInst20m~\cite{ziqiang2024marineinst} combines fifty data sources to yield a dataset with 19.2 million instance masks.

\subsubsection{Co-Occurrence With Advances in Self-Supervised Learning:}
Critically, this data revolution coincides with advances in self-supervised learning, which transform previously unusable unlabeled imagery into valuable training resources. Repositories such as CoralNet~\cite{beijbom2012automated}, XL CATLIN~\cite{gonzalez2014catlin, gonzalez2019seaview}, and Squidle+~\cite{squidle} contain millions of images with weak or sparse labels that are increasingly useful under modern frameworks (see Section~\ref{subsec:scene-understanding}). Platforms such as ReefCloud~\cite{AIMS2024ReefCloud, reefcloud}, CoralNet, and \citet{datamermaid} are also integrating semi-automated AI annotation tools, reducing cost and effort while expanding labeled datasets.  

The result is a positive feedback loop: better datasets enable stronger models, which attract more researchers and volunteers, leading to further data collection and methodological advances. This interplay of citizen science, CSR, gamification, specialized datasets, and self-supervised learning is fundamentally reshaping underwater AI and creating insights applicable to other domains where expert annotation is scarce and data collection is challenging.

\subsection{Scene Understanding}
\label{subsec:scene-understanding}

The underwater environment presents a unique combination of perceptual and operational challenges that have historically limited automated scene understanding. However, these same challenges now serve as catalysts for AI innovation, driving the development of novel techniques that advance not only marine robotics but the broader fields of computer vision and machine learning. Rather than viewing underwater conditions as obstacles to overcome, the AI community is increasingly recognizing them as opportunities to push methodological boundaries in ways that benefit numerous application domains. As a result, underwater scene understanding has undergone a fundamental transformation, shifting from narrow, task-specific models to general-purpose foundation models that can handle the full complexity of marine ecosystems. This evolution mirrors broader trends in AI but has been accelerated by the unique demands of underwater environments.

\begin{figure}[t]
    \centering
    \includegraphics[width=0.98\linewidth,clip,trim=4cm 4.5cm 4cm 4.5cm]{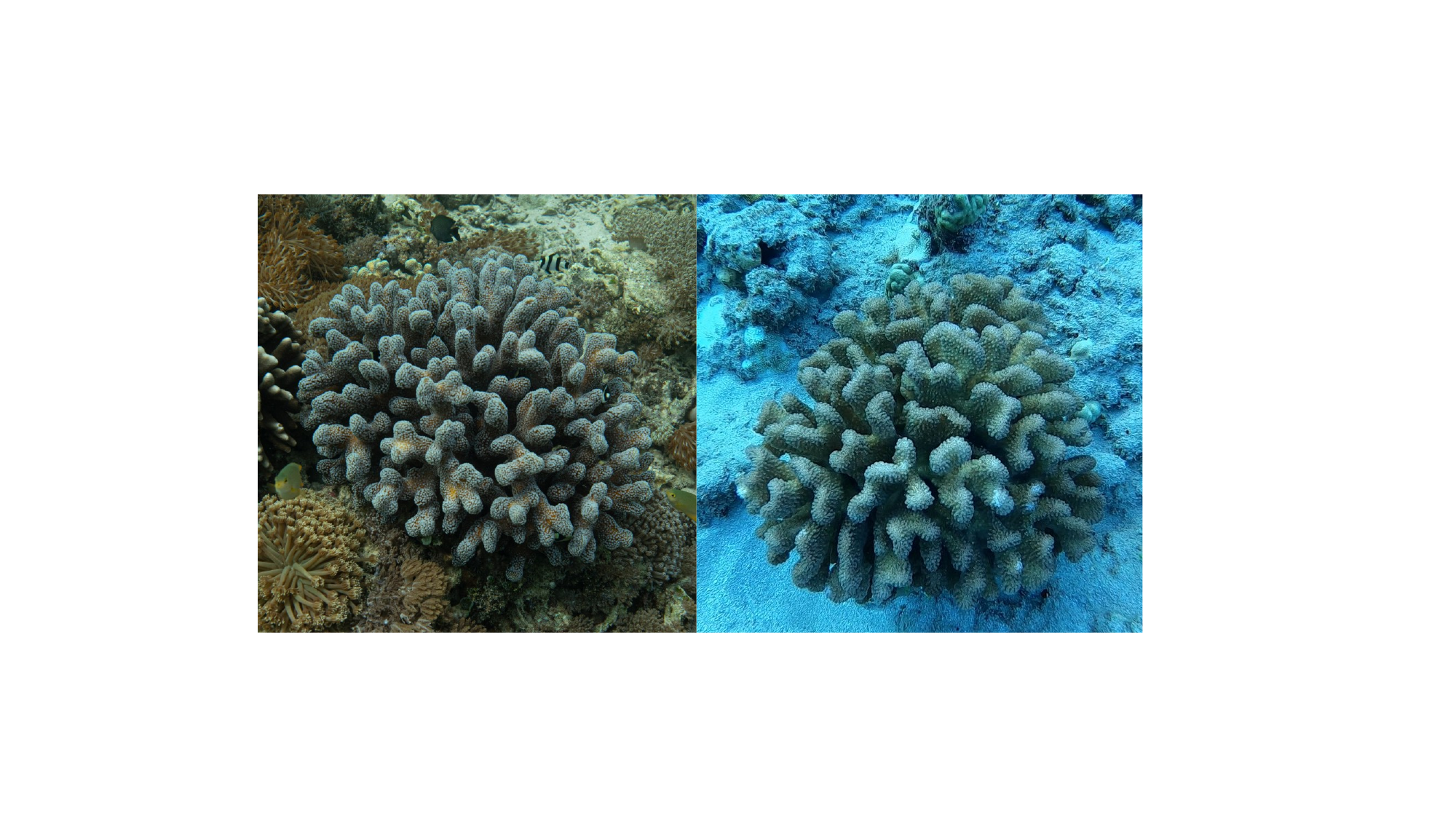}
    \caption{Fine-grained visual similarity of marine species.  Left: Coral from the \textit{Stylophora} family. Right: Coral from the \textit{Pocillopora} family. Images taken by Maria Costantini and Tobin Sparling and released under CC BY-NC.}
    \label{fig:coral}
\end{figure}

\subsubsection{Tackling Annotation Bottlenecks With Weakly Supervised Learning:}
Marine ecosystems exhibit extraordinary biodiversity, with over 800 hard coral species alone~\cite{dietzel2021population}, each requiring specialized taxonomic expertise to identify accurately (Fig.~\ref{fig:coral}). Traditional supervised learning approaches for marine species recognition rely on large, expertly-annotated datasets, a prohibitively expensive and slow process that creates a scalability bottleneck. This challenge has made the underwater domain a natural testbed for weakly supervised and few-shot learning approaches.

The marine biology community's existing practice of point-based sampling, born from the Coral Point Count method~\cite{kohler2006coral}, has given rise to recent advances in weakly supervised learning, creating a rich ecosystem of techniques that extract maximum information from minimal expert input. Point-based annotation in the underwater domain typically describes randomly distributed or grid-spaced sparse points which are labeled by marine scientists~\cite{alonso2019coralseg}.  This differs from standard computer vision domains, where it is common to have large datasets of images labeled with dense pixel-level labels.

Our recent point label aware superpixel method demonstrated that these sparse expert annotations can be effectively propagated by clustering pixels using learned features while respecting point label constraints, achieving substantial improvements while reducing computational requirements~\cite{raine2022point}. This success has inspired even weaker supervision approaches that operate with image-level labels alone, requiring 25 times fewer annotations than patch-level methods~\cite{raine2024image}. For seagrass meadow monitoring, contrastive pre-training combined with feature similarity has enabled coarse segmentation from image-level supervision, outperforming traditional patch-supervised methods on multi-species classification tasks~\cite{raine2024image}.

The emergence of foundation models has further transformed this landscape, with DINOv2 features now capable of generating multi-species coral masks without any domain-specific pre-training~\cite{raine2024human}. When integrated with human-in-the-loop labeling strategies that strategically leverage expert knowledge, these approaches achieve remarkable gains in efficiency~\cite{raine2024human}. Similarly, \citet{contini2025point} proposed a framework for weakly supervised semantic segmentation of coral reef aerial imagery, based on DINOv2 for generation of coarse spatial predictions and mask refinement with Segment Anything.  This progression from dense supervision to sparse points to image-level labels, and finally to foundation model-assisted annotation, exemplifies how domain constraints can drive methodological innovation with broad applicability across fields requiring expert knowledge integration.

\begin{figure}[t]
    \centering
    \includegraphics[width=\linewidth,clip,trim=3cm 7cm 2cm 4.5cm]{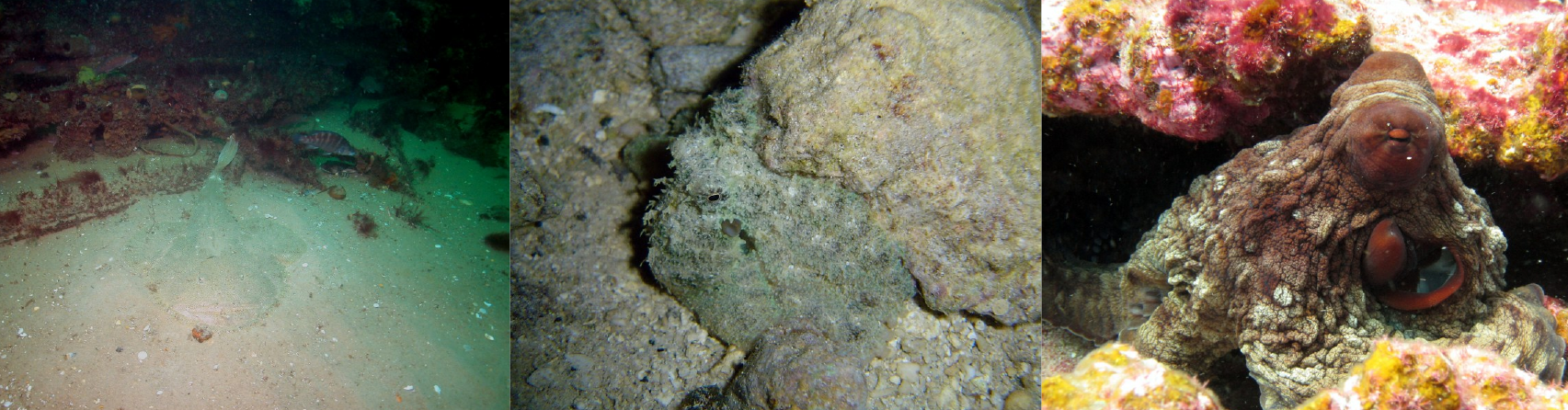}
    \caption{Cryptic marine species.  Left: a monkfish, center: a scorpionfish (image taken by David Burdick), and right: an octopus (image taken by Kevin Lino). Images are from the \citet{noaa} image collection, released into the public domain.}
    \label{fig:cryptic}
\end{figure}

\subsubsection{Visual Degradation as a Driver for Robust Perception:}
Underwater imagery suffers from systematic visual degradation that can cripple traditional computer vision systems: light attenuation causes dramatic color shifts with depth, suspended particles create backscatter and reduce contrast, and turbidity can render scenes nearly inscrutable. These conditions represent an extreme case of the domain shift and image corruption problems that plague real-world AI deployment across many applications.

There have been recent advances to tackle underwater visual degradation through transformer-based image enhancement~\cite{chen2024uwformer, qiao2025transformer}. Sauder et al. (2023) introduced self-supervised networks that estimate and correct backscatter and caustics from monocular sequences, improving real-time perception without ground-truth data. Similarly, \citet{wang2023underwater} and \citet{li2025spmformer} proposed physics-guided transformer architectures to restore contrast and correct color shifts in challenging underwater conditions.

\subsubsection{Addressing Cryptic Species and Camouflaging with Fine-Grained Recognition:}
Marine organisms have evolved sophisticated camouflage and cryptic coloration strategies that represent the ultimate challenge for visual recognition systems (Fig.~\ref{fig:cryptic}). Fish that perfectly mimic coral textures, octopuses that dynamically change color and pattern, and invertebrates that are virtually indistinguishable from their substrates push the boundaries of what current computer vision can achieve~\cite{hanlon2007cephalopod}.

Recent investigations into performance disparities across marine species have uncovered that not all detection challenges are created equal, and that data quantity alone cannot explain why certain species consistently underperform in automated detection systems. Analysis of underwater object detection reveals that foreground-background discrimination represents the most critical bottleneck for cryptic species like scallops, regardless of training data availability. Even when class imbalance is artificially corrected, persistent precision gaps indicate that certain marine organisms present intrinsic feature-based challenges that cannot be resolved through data augmentation alone~\cite{WilleAre}. This finding has profound implications for computer vision more broadly, suggesting that some visual recognition problems may be fundamentally algorithmic rather than data-driven.

\subsubsection{Foundation and Language Models in the Underwater Domain:}
The emergence of foundation models like CoralSCOP represents a paradigm shift from treating coral analysis as a collection of separate tasks (classification, detection, segmentation) to unified models capable of dense segmentation across multiple semantic granularities~\cite{zheng2024coralscop}. CoralSCOP addresses the fundamental challenge that coral structures are inherently amorphous and irregular—characteristics that generic segmentation methods fail to capture. Meanwhile, advances in self-supervised learning have enabled these models to leverage the vast quantities of unlabeled underwater imagery. CoralSRT~\cite{zheng2025coralsrt} demonstrated that foundation model features can be rectified through self-supervised guidance to achieve significant performance gains without additional human supervision.  

Recent work has seen forays into applying vision–language models and large language models to the underwater domain, though this remains an ongoing area of research. For example, MarineGPT demonstrated the integration of language models into marine applications and highlighted the potential for multimodal understanding, where visual perception can be combined with domain-specific knowledge expressed in natural language~\cite{zheng2023marinegpt}. Similarly, natural language prompting with the Contrastive Language-Image Pretraining (CLIP) model~\cite{radford2021learning} has been demonstrated for image pseudo-labeling, reducing annotation effort~\cite{raine2024image}. MarineInst~\cite{ziqiang2024marineinst} builds on these ideas and is a foundation model for marine imagery which outputs both instance masks and captions for object instances, and UWBench~\cite{zhang2025uwbench} proposes a comprehensive benchmark for underwater vision-language understanding.  These studies highlight the early potential for using foundation models in specialized domains. 

\subsection{Scene Reconstruction}
\label{subsec:3D}
Underwater scene reconstruction has undergone a fundamental paradigm shift: from traditional photogrammetry towards high resolution 3D scene rendering systems, capable of producing photo-realistic digital twins for precise ecosystem monitoring. This transformation represents a significant technical advance in marine robotics, with implications extending beyond underwater applications.

\subsubsection{Gaussian Splatting Influence:}

The emergence of underwater-specific 3D Gaussian Splatting methods has substantially improved reconstruction capabilities, with systems like WaterSplatting~\cite{li2024watersplatting}, SeaSplat~\cite{yang2024seasplat}, and UW-GS~\cite{wang2024uwgs} achieving real-time rendering while incorporating physics-based underwater image formation models. These approaches explicitly model light attenuation, backscatter, and wavelength-dependent absorption; moving beyond generic computer vision techniques toward domain-aware algorithms that account for underwater optics.

\subsubsection{Scale and Speed Transformation:}

Learning-based Structure from Motion has enabled reconstruction of kilometer-length reef surveys in minutes, a process that previously required hours of manual labor~\cite{sauder2024scalable}. These systems also integrate semantic segmentation directly into the reconstruction pipeline, unifying geometric mapping and scene understanding. Recent advances in multi-view stereo combined with physical imaging models~\cite{fast2025underwater} further accelerate processing while maintaining geometric accuracy. This represents a shift from post-hoc analysis to rapid, in-situ ecosystem monitoring capabilities that can support active marine conservation efforts.

\subsubsection{Integration with Marine Science Workflows:}

The coupling of 3D reconstruction with semantic segmentation has progressed from technical demonstrations to operational marine science tools. Multi-modal approaches combining geometric reconstruction with automated species classification~\cite{zhong2023combining,remmers2025rapidbenthos} enable direct computation of ecological metrics, transforming how marine biologists approach large-scale surveys and longitudinal studies.  \citet{GorryIROS2025} have demonstrated that visual place recognition can be integrated with feature matching and image segmentation to enable robust identification of revisited underwater areas for ecosystem change analysis. These integrated systems demonstrate how AI-driven reconstruction can simultaneously provide geometric understanding and biological insights at scales previously impractical with traditional survey methods.

\subsubsection{Digital Twin Emergence:}
Underwater reconstruction has evolved toward ecosystem-scale digital twins with demonstrated deployments spanning multi-million dollar research programs~\cite{whoi2024digital} and into the tens to hundreds of hectares~\cite{KAUST}. These systems enable virtual experimentation with management interventions and provide the foundation for predictive ecosystem modeling at unprecedented scales. These underwater reconstruction systems comprise dynamic, semantically aware models which support active marine conservation and timely ecosystem management.

This evolution from passive documentation to active environmental intelligence represents a broader trend in AI systems moving from observation to intervention, with underwater applications serving as a proving ground for techniques that will likely find application across numerous challenging perception domains.

\subsection{From Research to Impact}
\label{subsec:applications}

The emerging trend of AI for underwater environments has already begun to deliver tangible benefits for fisheries, aquaculture, and reef restoration. What was once primarily a research endeavor has transitioned into a suite of operational tools with real-world impact.

\subsubsection{Fisheries:} AI-powered monitoring systems are transforming fishery management with greater transparency, efficiency, and accountability. Deep learning applied to underwater imagery and sonar now enables automated fish counting and biomass estimation, supporting sustainable stock assessments at scale~\cite{connolly2022fish,connolly2023out}. In parallel, AI fused with satellite and Automatic Identification System data is revolutionizing compliance monitoring, with novel approaches detecting illegal ``dark'' fishing vessels that disable trackers~\cite{raynor2025little}.

\subsubsection{Aquaculture:} Commercial aquaculture has increasingly adopted AI-enabled Internet-of-Things (AIoT) platforms for operational efficiency. Smart feeding systems combine underwater cameras with behavioral analysis to optimize feed delivery, improving growth while reducing waste~\cite{tidal}. Additional developments include water quality monitoring, health diagnostics (e.g., disease detection), and real-time biomass estimation~\cite{connolly2022fish, singh2022sustainable}, greatly reducing labor requirements and costs. Recent surveys highlight that AIoT applications now span feeding, water chemistry, fish health, and even individual animal tracking~\cite{ubina2023digital, hu2023design}.

\subsubsection{Reef Restoration:} Robotics and AI are scaling coral propagation and rehabilitation efforts. A notable example is the Coral Growout Robotic Assessment System (CGRAS), which uses robotic imaging to automatically detect and count millions of baby corals in hatchery tanks~\cite{cgras}. This replaces thousands of hours of manual labor, providing scientists with real-time growth data and enabling large-scale coral deployment across the Great Barrier Reef. 

Complementary initiatives, such as LarvalBots~\cite{mou2022reconfigurable} for targeted larval dispersal and Reef Guidance Systems (RGS) for site selection, further accelerate reef gardening and restoration~\cite{raine2025aidriven}.  The impact of restoring reefs can be established using AI with underwater cameras to quantify fish stock recovery~\cite{mcaneney2025fish, connolly2024estimating}.   The ReefOS system~\cite{coralgardeners} combines static cameras, fish detection models, a custom photogrammetry toolkit, and a geospatial visualization dashboard to perform fine-scale monitoring of their restoration sites in French Polynesia, Fiji and Thailand. 

\subsubsection{Bridging AI Outputs to Ecological Metrics:}
To ensure uptake, AI outputs are increasingly being mapped onto conventional monitoring protocols (e.g., percent cover from point-intercept). Platforms like ReefCloud and CoralNet operationalize random point sampling and class predictions to yield comparable ecological indicators and dashboards~\cite{AIMS2024ReefCloud, beijbom2012automated}. For 3D products, standardized co-registration and uncertainty budgets allow statistically defensible temporal change estimates that align with regulatory expectations \cite{lechene2024evaluating}.

\subsubsection{Generalization, Adaptation, and Uncertainty:}
Generalization across regions and sensors remains central. Broad, curated image corpora (e.g., FathomNet) and new multi-region benthic datasets (e.g., BenthicNet) provide the diversity needed for pretraining, open-set handling, and rapid adaptation \cite{katija2022fathomnet,sauder2025coralscapes, lowe2025benthicnet}. Early works into out-of-distribution detection~\cite{wyatt2025safe} are supporting safer deployment of AI algorithms.  In parallel, uncertainty-aware evaluation (model confidence, inter-observer variability, and protocol variance) is increasingly reported alongside accuracy, supporting use in regulation and markets (e.g., blue carbon monitoring, reporting and validation; marine protected area compliance).  

\section{Conclusions}
\label{sec:conclusion}

AI for underwater perception illustrates a bidirectional relationship between environmental necessity and technological innovation. Conservation needs have accelerated research, while underwater constraints have driven advances in computer vision and machine learning that benefit many domains. This dual impact demonstrates that domain-specific engagement can drive both fundamental research and tangible impact.

The convergent drivers, i.e.~environmental imperative, data democratization, and research opportunity, have created a self-reinforcing ecosystem where new methods enable applications, generate data, and attract investment—transforming underwater AI into a recognized research frontier with real-world impact. Core challenges such as turbidity, cryptic species detection, and annotation bottlenecks have spurred innovations in weak supervision, open-set recognition, and robust perception in degraded environments. These advances extend beyond marine science to applications like medical imaging, analysis of satellite data, and autonomous systems.

The field exemplifies how AI research can progress from academic novelty to operational capability. Multi-million-dollar deployments in reef monitoring and restoration provide a template for translating innovation into measurable societal outcomes. Underwater AI also serves as a proving ground for emerging techniques in foundation models and human–AI collaboration. The requirements for uncertainty quantification, expert knowledge integration, and robust performance under uncertainty make it an ideal domain to mature reliable AI systems for high-stakes environments.

\section*{Acknowledgments}
S.R.~and T.F.~acknowledge continued support from the Queensland University of Technology (QUT) through the Centre for Robotics. T.F.~acknowledges funding from an Australian Research Council (ARC) Discovery Early Career Researcher Award (DECRA) Fellowship DE240100149. S.R. and T.F.~ acknowledge support from the Reef Restoration and Adaptation Program (RRAP) which is funded by a partnership between the Australian Government’s Reef Trust and the Great Barrier Reef Foundation.

\bibliography{aaai2026}

\end{document}